\documentclass{article}

\usepackage{arxiv}
\usepackage[utf8]{inputenc} 
\usepackage[T1]{fontenc}    
\usepackage{hyperref}       
\usepackage{url}            
\usepackage{booktabs}       
\usepackage{amsfonts}       
\usepackage{nicefrac}       
\usepackage{microtype}      
\usepackage{lipsum}
\usepackage{tabularx}
\usepackage{array}
\usepackage{booktabs}
\usepackage{multirow}
\usepackage{makecell}
\usepackage{graphicx}
\usepackage{natbib}
\graphicspath{ {./images/} }

\title{age-diverse deepfake dataset: bridging the age gap in deepfake detection}

\renewcommand{\headeright}{}
\renewcommand{\undertitle}{}
\renewcommand{\shorttitle}{Age-Diverse Deepfake Dataset: Bridging the Age Gap in Deepfake Detection}

\author{
Unisha Joshi \vspace{0.8 em}\\
College of Engineering and Technology\\
Grand Canyon University,USA\\
\texttt{ujoshi@my.gcu.edu} \\
}
\makeatletter
\renewcommand{\@date}{}
\makeatother

\begin{document}
\maketitle

\begin{abstract}
The challenges associated with deepfake detection are increasing significantly with the latest advancements in technology and the growing popularity of deepfake videos and images. Despite the presence of numerous detection models, demographic bias in the deepfake dataset remains largely unaddressed. This paper focuses on the mitigation of age-specific bias in the deepfake dataset by introducing an age-diverse deepfake dataset that will improve fairness across age groups. The dataset is constructed through a modular pipeline incorporating the existing deepfake datasets Celeb-DF, FaceForensics++, and UTKFace datasets, and the creation of synthetic data to fill the age distribution gaps. The effectiveness and generalizability of this dataset are evaluated  using three deepfake detection models: XceptionNet, EfficientNet, and LipForensics. Evaluation metrics, including AUC, pAUC, and EER, revealed that models trained on the age-diverse dataset demonstrated fairer performance across age groups, improved overall accuracy, and higher generalization across datasets. This study contributes a reproducible, fairness-aware deepfake dataset and model pipeline that can serve as a foundation for future research in fairer deepfake detection. The complete dataset and implementation code are available at \href{https://github.com/unishajoshi/age-diverse-deepfake-detection}{https://github.com/unishajoshi/age-diverse-deepfake-detection}.

\vspace{1 em}

\end{abstract}

\section{Introduction}
Deepfake technology has emerged as a powerful outcome of deep learning and synthetic media generation, rapidly transforming digital content creation with significant applications in entertainment, education, and virtual experiences. However, it also poses a significant risk of being exploited by malicious actors using them to mislead audiences, causing a significant threat to society and politics \cite{nadimpalli2022gbdfgenderbalanceddeepfake}. 

While different deepfake detection methods have been introduced to counter these threats, the publicly available datasets that are used in deepfake detection models, such as FaceForensics++\cite{rossler2019faceforensics++}, and Celeb-DF \cite{li2020celebdflargescalechallengingdataset} lack clarity about the fairness of the dataset for demographic parameters such as age, gender, and ethnicity \cite{nadimpalli2022gbdfgenderbalanceddeepfake} . The lack of age diversity in the deepfake dataset impacts the deepfake detection model and its performance, potentially compromising the fairness and trustworthiness of this technology in future \cite{xu2024analyzingfairnessdeepfakedetection}.

Analysis of age diversity and creation of a fairer deepfake dataset can help to address this critical gap. This paper aims to assess the fairness of deep-fake datasets specific to age demographics, with the central focus of developing a comprehensive, age-diverse deep-fake dataset including a wide range of age groups from teens to elderly individuals. The generated dataset has the goal of enhancing the fairness and accuracy of the deep-fake detection system in comparison to the source dataset, while also addressing the age gap present in them. 

The main contributions of this paper include several key achievements towards reducing age bias in deepfake detection. Firstly, the study provides proper age level annotations for the widely used publicly available datasets FaceForensics++ \cite{rossler2019faceforensics++} and Celeb-DF \cite{li2020celebdflargescalechallengingdataset}. The analysis of fairness in these datasets is conducted with the proper age annotations and conducting an exploratory evaluation of age distribution using different age groups, providing insight into the need for augmentation in underrepresented age groups.
For age groups with underrepresented records, synthetic data are generated using InsightFace \cite{deng2018menpo} \cite {guo2021sample} \cite {deng2020subcenter} and SimSwap\cite{10.1145/3394171.3413630} technology to address the gap in age diversity.  The resulting dataset combines data from multiple sources with newly created synthetic data to produce a comprehensive, annotated, age-diverse deepfake dataset. 

To evaluate the effectiveness of the resulting dataset, deepfake detection models, specifically XceptionNet \cite{chollet2017xceptiondeeplearningdepthwise},  EfficientNet \cite{Koonce2021}, and LipForensics \cite{haliassos2021lipsdontliegeneralisable} are employed. The performance of these models are assessed using standard metrics such as AUC, pAUC and EER. Finally, this paper presents the comparative performance analysis between the final dataset and the source datasets using deepfake detection models, along with an age-specific performance evaluation to assess the effectiveness of the dataset across different age groups and the impact of age diversity on model performance and fairness.

\section{Related Works}
\subsection{Fairness Analysis of Deepfake Datasets}
Deepfake datasets that are publicly accessible do not have sufficient annotations to be utilized in the deepfake detection process or to examine the impact of demographic factors \cite{xu2024analyzingfairnessdeepfakedetection}. An extensive fairness evaluation conducted in \cite{xu2024analyzingfairnessdeepfakedetection} addressed this gap with proper annotations to multiple popular datasets, including FaceForensics++, Celeb-DF, DFDC, DeeperForensics-1.0, and DFD, analyzing the performance across various demographic aspects such as gender, age, hair color, skin color, and facial attributes. The overall analysis revealed data disparity across multiple demographics. The evaluation performed using EfficientNet-B0, Xception, and Capsule-Forensics-V2 revealed bias in performance, hindering the models’ learning patterns. In addition to that, models trained on demographically skewed datasets have lower performance scores for underrepresented groups despite higher overall accuracy \cite{ju2023improving}.

These studies highlight the impact of demographic imbalance on model performance and the occurrence of unfair outcomes in deepfake detection. These insights collectively highlight the necessity for balanced demographic representation in deepfake detection, reinforcing the motivation of this paper to improve age fairness in deepfake detection.

\subsection{Analysis of Gender Demographics}
An analysis of existing deepfake datasets reveals a significant gender imbalance and the presence of inconsistent face swaps \cite{nadimpalli2022gbdfgenderbalanceddeepfake}. The GBDF dataset addresses this issue by manually annotating gender in the FaceForensics++ and Celeb-DF, resulting in improved fairness across gender demographics \cite{nadimpalli2022gbdfgenderbalanceddeepfake}. The evaluation using detection models showed a significant bias in the original datasets, achieving higher performance on the male group compared to the female group. However, the model trained with the gender balanced dataset led to more impartial performance across genders without impacting overall detection accuracy. This work highlights a broader concern regarding demographic bias in deepfake detection and serves as a foundation for this study’s focus on enhancing age diversity and fairness.

\section{Methodology}
This section presents the step-by-step process implemented while building an age-diverse deepfake dataset and evaluates its influence on detection model performance and fairness. The methodology follows a structured pipeline, as illustrated in Figure 1, beginning with importing existing deepfake datasets, followed by extracting frames and annotating them with age information. To address the lack of age representation in certain age groups, synthetic videos are generated and combined with the dataset. This enriched dataset is used to train multiple deepfake detection models, followed by an evaluation process. During the evaluation, overall performance, age-specific performance, and cross-source performance are captured and analyzed. 
\begin{figure}[htbp]
    \centering
    \includegraphics[width=0.95\linewidth]{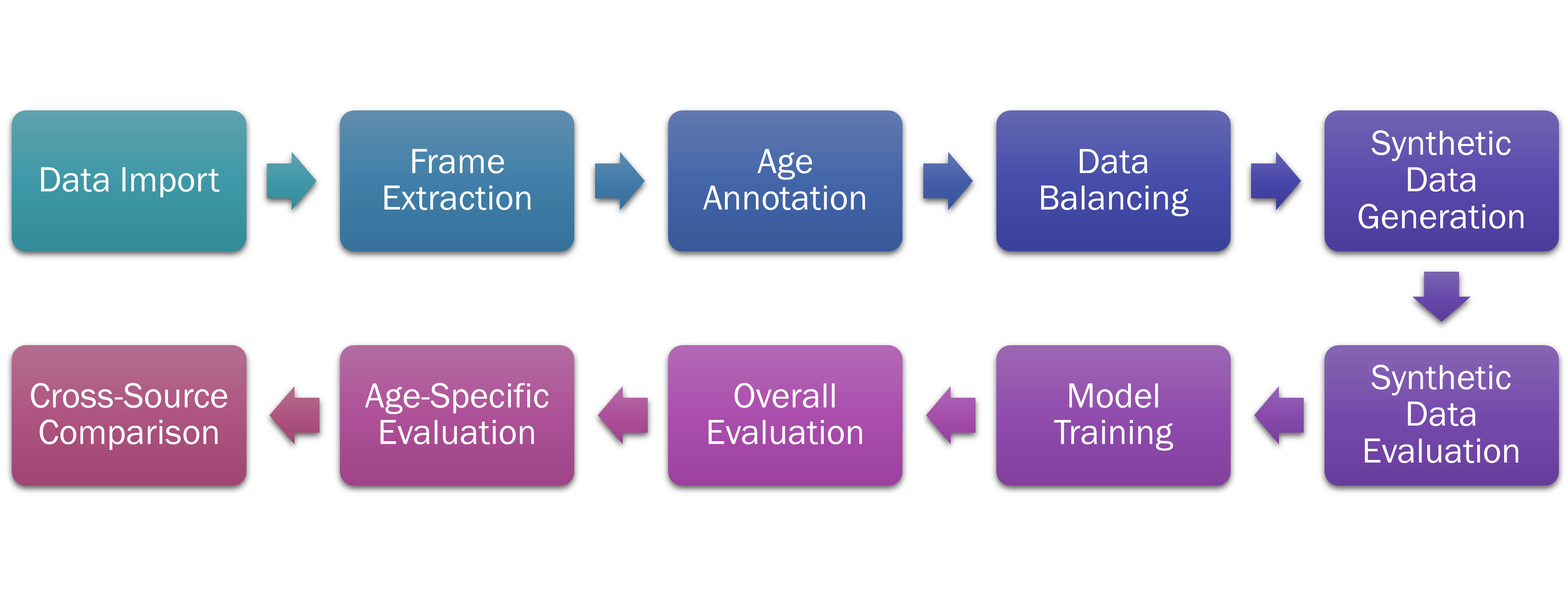}
    \caption{\textit{Workflow Pipeline for Age-Diverse Deepfake Detection.}}
    \label{fig:method_pipeline}
\end{figure}
\subsection{Understanding Data}
This project incorporates three publicly available datasets: Celeb-DF \cite{li2020celebdflargescalechallengingdataset}, FaceForensics++ \cite{rossler2019faceforensics++}, and UTKFace [13]. The Celeb-DF deepfake dataset contains 590 real and 5636 fake high-resolution videos. Celeb-DF utilizes an encoder-decoder model, generating videos with high visual quality and minimizing artifacts like splicing borders, color mismatch, and face orientation inconsistencies \cite{nadimpalli2022gbdfgenderbalanceddeepfake}. In this paper, all 590 real videos and subsets of 537 fake videos include facial swaps for each actor in real videos. This selection strategy is implemented to mitigate the overrepresentation of young and adult age groups in the final dataset. The FaceForensics++ dataset is used for detecting facial manipulations and consists of deepfake videos created using techniques such as Identity Swapping (FaceSwap, FaceSwap-Kowalski, FaceShifter, DeepFakes) and Expression Swapping (Face2Face and Neural textures) \cite{nadimpalli2022gbdfgenderbalanceddeepfake}. From this dataset, all 1001 real videos and 1000 fake videos generated specifically using the FaceSwap technique are used in this study. In contrast, UTKFace has 20852 real facial images with proper demographic labels for age, gender, and ethnicity \cite{zhifei2017cvpr}. This dataset serves as the foundation for age-based synthetic data augmentation, combining with other source videos to generate age-diverse deepfakes.

To perform age-specific analysis and further processing, 30 evenly spaced frames were extracted from each video in the Celeb-DF and FaceForensics++ dataset using OpenCV’s cv2.VideoCaptre module \cite{bradski2000opencv}. This approach ensures frame selection diversity across the video duration and restricts data redundancy. Each frame was preserved in its original resolution and encoded in JPEG format for consistency. To illustrate the nature of extracted frames, Figure 2 presents visual examples from both real and fake videos.

\begin{figure}[htbp]
    \centering
    \includegraphics[width=0.95\linewidth]{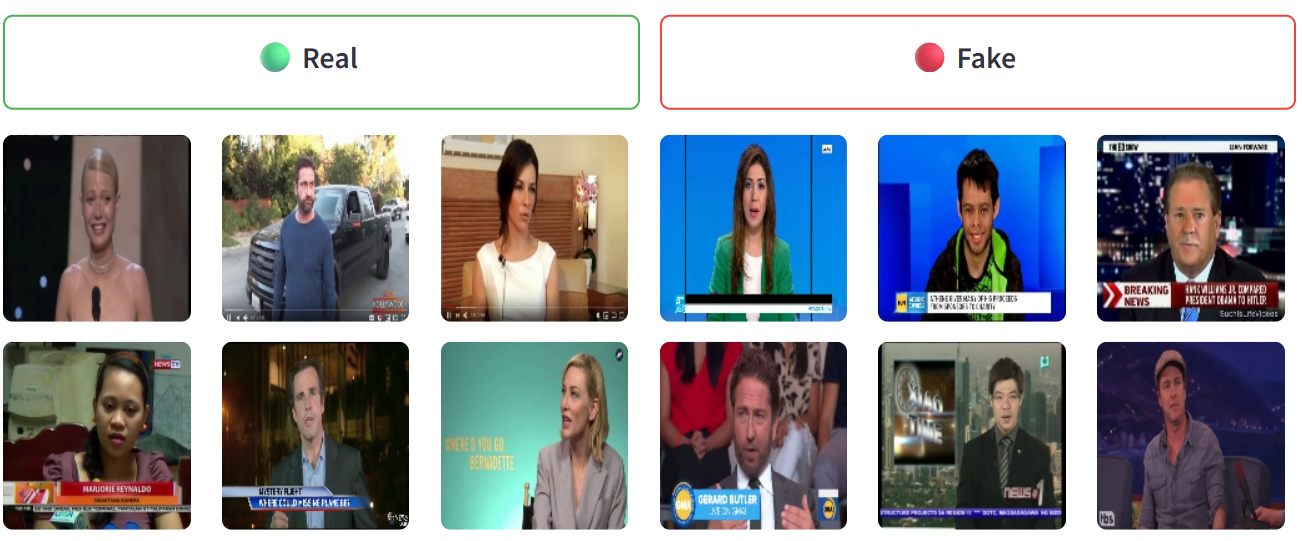}
    \caption{\textit{Real and Fake Frames Visualization of data from Celeb-DF and FaceForensics++ datasets}}
    \label{fig:real_and_fake_frames}
\end{figure}

information for filename, label (real or fake), source dataset, estimated age, and age group, then compiled into a structured annotation file.

\begin{table}[ht]
\centering
\caption{Age and Label Distribution by Dataset}
\renewcommand{\arraystretch}{1.3}
\begin{tabular}{|l|r|r|r|}
\hline
\textbf{Label \& Age Group} & \textbf{UTKFace} & \textbf{Celeb} & \textbf{FaceForensics++} \\
\hline
fake (19--35)    & 0     & 427   & 802 \\
\hline
fake (36--50)    & 0     & 108   & 191 \\
\hline
real (0--10)     & 2373  & 0     & 0   \\
\hline
real (10--18)    & 1351  & 0     & 0   \\
\hline
real (19--35)    & 10248 & 463   & 792 \\
\hline
real (36--50)    & 3860  & 122   & 193 \\
\hline
real (51+)       & 4416  & 1     & 4   \\
\hline
\end{tabular}
\end{table}
\begin{figure}[htbp]
    \centering
    \includegraphics[width=0.95\linewidth]{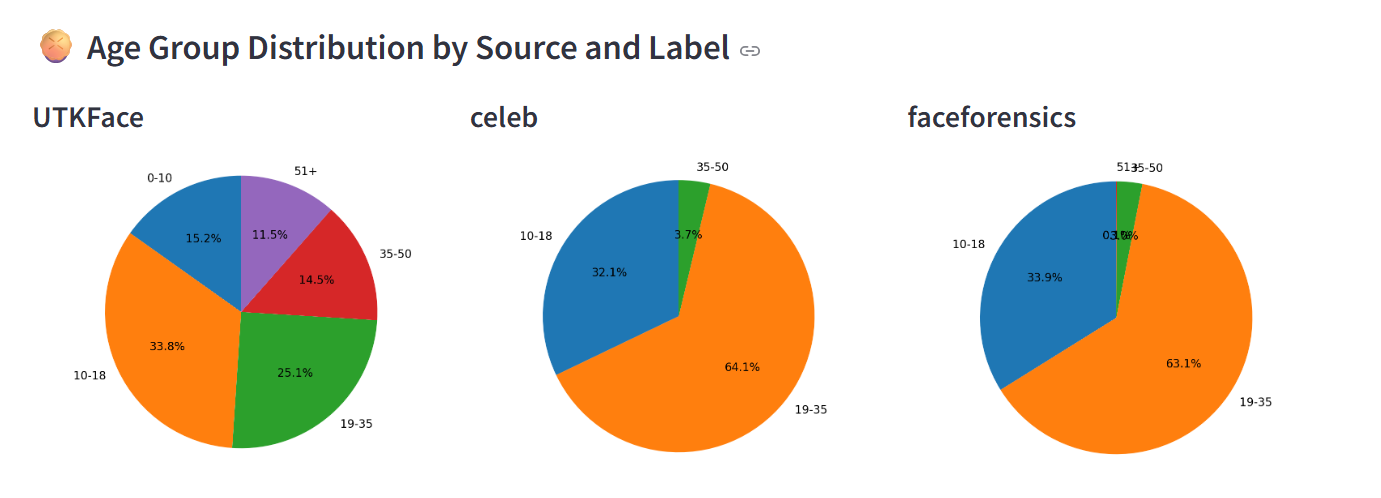}
    \caption{\textit{Age group distribution pie chart by Source Dataset}}
    \label{fig:age_label_distribution}
\end{figure}
Following extraction, age annotations were applied using DeepFace Framework \cite{serengil2024lightface}, which internally employs a face detector and the VGG-Face model \cite{parkhi2015deep} for face representation and age estimation. Each extracted frame was passed through an annotation process to detect the face and estimate its age. Frames lacking a detectable face were skipped during annotation and removed from further processing, while frames containing multiple faces were processed using only the most prominent detected face to maintain consistency and annotation quality. The estimated age for each frame was binned into groups: 0-10, 10-18, 19-35, 36-50, and 51+, representing age demographics from child to elderly. The final annotation was stored, including metadata. 
The age-label distribution, summarized in Table 1, reveals a significant demographic skew in the source dataset, with a heavy concentration of samples in the 19-35 age group and minimal representation for 10-18, 36-50, and 51+ brackets. This imbalance is visually supported by Figure 3, which provides a pie chart representation of age group proportions across the datasets. This figure also highlights the age diversity present in the UTKFace dataset, demonstrating its potential as a suitable source for generating synthetic data to address age imbalance in the overall dataset. The age disparity in the Celeb-DF and Faceforensics++ dataset stresses the need for targeted data augmentation to uplift age diversity, which is essential for improving the fairness and robustness of deepfake detection models. 

\subsection{Video Augmentation for Age Diversity}
To address the imbalance of age diversity in the source datasets (FaceForensics++ \cite{rossler2019faceforensics++}, and Celeb-DF \cite{li2020celebdflargescalechallengingdataset}), a structured augmentation strategy was implemented. Firstly, the undersampling was performed to reduce the counts for the majority age groups. For this, the mean frame count across all age groups was computed, including data from both Celeb-DF and FaceForensics++ datasets, then the age groups with records exceeding the mean value were reduced to match the mean count. To mitigate the shortage of samples in real age brackets, real fake images from the UTKFace dataset were embedded in the dataset. Table 2 displays the data counts after undersampling and prior to synthetic data generation, which was used to create the synthetic data generation plan and determine the number of required synthetic video files. After that, the synthetic data generation process is performed to fill the gaps in the fake age groups. 
\renewcommand{\arraystretch}{1.3} 

\begin{table}[ht]
\centering
\caption{Dataset after performing undersampling of the majority age groups}
\begin{tabular}{|l|r|r|r|}
\hline
\textbf{Label \& Age Group} & \textbf{UTKFace} & \textbf{Celeb} & \textbf{FaceForensics++} \\
\hline
fake (19--35)    & 0     & 254   & 510 \\
\hline
fake (36--50)    & 0     & 108   & 191 \\
\hline
real (0--10)     & 525   & 0     & 0   \\
\hline
real (10--18)    & 525   & 0     & 0   \\
\hline
real (19--35)    & 0     & 192   & 333 \\
\hline
real (36--50)    & 210   & 122   & 193 \\
\hline
real (51+)       & 520   & 1     & 4   \\
\hline
\end{tabular}
\end{table}

The augmentation was performed using the SimSwap framework \cite{10.1145/3394171.3413630}, which is a face-swapping model based on an encoder-decoder architecture. This model retains the identity features from the source image while transforming them into the target video, preserving the target’s pose and expression. This approach allows for the generation of realistic deepfake content required for underrepresented age groups.

To enhance the quality and relevance of the generated videos, Insightface \cite{deng2018menpo} \cite {guo2021sample} \cite {deng2020subcenter} was integrated into the synthetic video generation pipeline for advanced facial feature extraction and similarity matching. InsightFace \cite{deng2018menpo} \cite {guo2021sample} \cite {deng2020subcenter} offers robust face embeddings and facial attributes such as yaw, pitch, brightness, and expression using ResNet-100 \cite{he2015deepresiduallearningimage} . To ensure selection of the best match, each real image from UTKFace dataset \cite{zhifei2017cvpr} was matched with a target video (from Celeb-DF or FaceForensics++) by computing cosine similarity between them and matching facial attributes (specifically yaw, pitch, expression and brightness) scores. 

After the best match section using Insightface, SimSwap was implemented to produce synthetic videos swapping the real UTKFace source face into the target video frames. To ensure the quality of synthetic videos, the Structural Similarity Index Measure (SSIM) \cite{1284395}, and the Peak Signal-to-Noise Ratio (PSNR) were calculated for each generated video. Overall and real-time SSIM and PSNR scores were monitored during each swap operation to detect anomalies or failures in generation. The average SSIM
score across all synthetic videos was 0.4053, indicating moderate structural similarity. The average PSNR score was 0.28 dB, indicating good visual quality with low distortion. 

The resulting dataset derived after the synthetic data generation combines original real and fake videos with newly generated synthetic data for underrepresented age groups. The final age-diverse dataset, as illustrated in Table 3, reflects a balanced distribution of real and fake videos across all age groups (0-10, 10-18, 19-35, 36-50, 51+). In case of real videos, each age group contains exactly 525 videos, ensuring equal representation across the real category. For fake videos, the video count ranges between 697 to 764, which was due to the availability of real and face matches, facial similarity score, and match quality score filters. However, this minor variation is not expected to show any significant impact on the model performance as the dataset remains largely balanced in terms of age distribution.

\begin{table}[ht]
\centering\begin{minipage}{0.8\textwidth} 

\caption{Final Age-diverse Dataset Prepared, combining Celeb-DF, FaceForensics++, and Synthetic Videos for Training}
\begin{tabularx}{\linewidth}{|>{\centering\arraybackslash}X|>{\centering\arraybackslash}X|>{\centering\arraybackslash}X|}
\hline
\textbf{Age Group} & \textbf{Fake} & \textbf{Real} \\
\hline
0--10   & 743 & 525 \\
\hline
10--18  & 697 & 525 \\
\hline
19--35  & 764 & 525 \\
\hline
36--50  & 753 & 525 \\
\hline
51+     & 745 & 525 \\
\hline
\end{tabularx}
\end{minipage}
\end{table}

\subsection{Model Training}
To assess the effectiveness of the resulting age-diverse dataset, three deepfake detection models (XceptionNet \cite{chollet2017xceptiondeeplearningdepthwise}, EfficientNet-B0 \cite{Koonce2021}, and LipForensics \cite{haliassos2021lipsdontliegeneralisable}) were selected. 

\begin{itemize}
    \item XceptionNet, based on depthwise separable Convolutional Neural Network (CNN), effectively extracts detailed visual features, making it suitable for detecting crucial artifacts in deepfake videos \cite{10423094}.
    \item EfficientNet-B0 model implements a compound scaling strategy to achieve high accuracy with low computational cost, making it suitable for large-scale video frame classification \cite{Koonce2021}. 
    \item LipForensics, based on the ResNet18 \cite{he2015deepresiduallearningimage} architecture, involves the pipeline of cropping and aligning the lip region, using temporal modeling, and training using the ResNet18 model \cite{haliassos2021lipsdontliegeneralisable}. 
\end{itemize}

 These models were selected for their proven effectiveness in deepfake detection and their architecture diversity supporting a more robust comparative analysis. All three models (XceptionNet, EfficientNet, and LipForensics) were trained independently at the frame level to maximize the data availability and capture underlying inconsistency. 
 
 The complete model training process is visually summarized in Figure 4, illustrating the sequential model training stages from data preparation to trained model output.
\begin{figure}[htbp]
    \centering
    \includegraphics[width=0.95\linewidth]{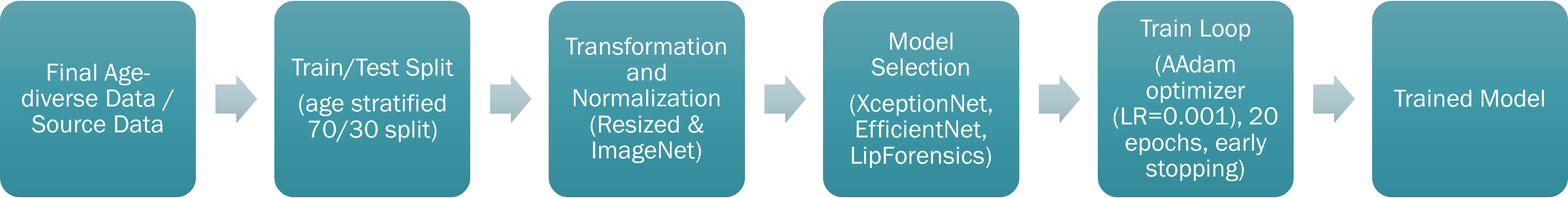}
    \caption{\textit{Deepfake Detection Model Pipeline}}
    \label{fig:model_pipeline}
\end{figure}

Prior to training, the final age-diverse dataset and source datasets were stratified into a 70:30 train-test split, preserving the proportion distribution of age group and class labels (real/fake). Each input frame was resized to 224X224 and normalized using ImageNet \cite{10.1145/3065386} Statistics to ensure compatibility with pre-trained model backbones. Training was conducted using the Adam optimizer \cite{kingma2017adammethodstochasticoptimization} with a learning rate of 0.001 and a weight decay of 1e-6 has been utilized. Each model was trained for 20 epochs with a batch size of 32, and early stopping was employed based on the validation AUC (Area Under the Curve) with a patience of 3 epochs to prevent overfitting.

The loss function used was Binary Cross-Entropy with Logits Loss (BCEWithLogitsLoss) \cite{ruby2020binary},which combines a Sigmoid activation and the Binary Cross-Entropy (BCE) loss in a single step, enabling models to perform binary classification directly on raw logits, improving numerical stability during the training process [24].This training configuration ensures consistent evaluation across models and datasets, enabling reliable assessments of performance and fairness.

\subsection{Evaluation Strategy}
To evaluate the performance and fairness, trained models were evaluated using three different datasets: a generated age-diverse dataset, the original Celeb-DF \cite{li2020celebdflargescalechallengingdataset} dataset, and the original FaceForensics++ \cite{rossler2019faceforensics++} dataset. For each model, performance difference for standard evaluation metrics (AUC, pAUC, and EER) was calculated. The AUC (Area Under the ROC Curve) measures the overall classification performance, pAUC (Partial AUC) accesses the perfromace at low false-positive rate, and the EER (Equal Error Rate) indicates the threshold where false positive and false negative rates are equal, assessing the model accuracy.

The evaluation will be performed in three stages: overall evaluation, age-specific evaluation and cross-dataset comparison. The goal of overall evaluation will be to assess the model performance on overall dataset without considering the age distribution. In case of age-specific evaluation, it will includes a more detailed analysis of model performance across different age group labels. Finally, in cross data comparison, the performance of model trained with final age-diverse dataset will be compared against the model trained with source (Celeb-DF and FaceForensics++) datasets. 

This strategy is implemented to cover overall classification accuracy, fairness across age demographics, and generalization across datasets.

\subsection{Findings}  
This section presents a comparative evaluation of deepfake detection models (XceptionNet \cite{chollet2017xceptiondeeplearningdepthwise}, EfficientNet \cite{Koonce2021}, and LipForensics \cite{haliassos2021lipsdontliegeneralisable}), where these models were validated against the test set from the age diverse dataset, original Celeb-DF \cite{li2020celebdflargescalechallengingdataset} set, and the FaceForensics++ \cite{rossler2019faceforensics++} dataset. All the evaluation metrics and their findings are discussed below:
\subsubsection{Overall Evaluation}
The primary goal of this evaluation is to assess model performance on the overall dataset without considering the age group distribution. For this evaluation, model trained with generated age-diverse datasets are evaluated using the test datasets from all three datasets (age-diverse dataset, Celeb-DF dataset \cite{li2020celebdflargescalechallengingdataset} , and Faceforensics++ \cite{rossler2019faceforensics++} dataset). This evaluation will provide a high-level view of effectiveness of age-diverse dataset for improving the overall model performance irrespective of selected dataset.

\begin{table}[ht]
\renewcommand{\arraystretch}{1.3}
\centering
\caption{Age-diverse dataset model performance evaluation across datasets}
\begin{tabular}{|l|c|c|c|c|c|c|c|c|c|}
\hline
\textbf{Test Set} & \multicolumn{3}{c|}{\textbf{XceptionNet}} & \multicolumn{3}{c|}{\textbf{EfficientNet}} & \multicolumn{3}{c|}{\textbf{LipForensics}} \\
\hline
& AUC & PAUC & EER & AUC & PAUC & EER & AUC & PAUC & EER \\
\hline
Age-Diverse         & 0.9983 & 0.9993 & 0.0134 & 0.9994 & 0.9998 & 0.0084 & 0.9970 & 0.9990 & 0.0213 \\
\hline
Celeb-DF           & 0.9963 & 0.9954 & 0.0214 & 0.9985 & 0.9983 & 0.0146 & 0.9927 & 0.9927 & 0.0365 \\
\hline
FaceForensics++  & 0.9992 & 0.9992 & 0.0085 & 0.9997 & 0.9997 & 0.0056 & 0.9977 & 0.9981 & 0.0202 \\
\hline
\end{tabular}
\end{table}

Table 4 illustrates the performance of models across all datasets. For the age-diverse dataset, all three models achieved the AUC and PAUC values higher than 0.997, indicating high accuracy of distinguishing real from fake data. The EfficientNet \cite{Koonce2021} model has the highest AUC of 0.9994, PAUC of 0.9998, and the lowest EER of 0.0084, indicating excellent generalization. XceptionNet \cite{chollet2017xceptiondeeplearningdepthwise} model also has a strong performance with an AUC of 0.9983, PAUC of 0.9993, and an EER of 0.0134. Among the three models, LipForensics \cite{haliassos2021lipsdontliegeneralisable} model has slightly lower scores but still strong performance with values of AUC of 0.997, PAUC of 0.999, and EER of 0.0213. 

In the case of testing with the Celeb-DF dataset, XceptionNet maintained a strong performance with an AUC score of 0.9963, and a slightly reduced score for EfficientNet with an AUC score of 0.9985. For LipForensics, the performance scores have reduced noticeably with scores of AUC of 0.9927, PAUC of 0.9927, and EER of 0.0365. 

Now, when tested with the FaceForensics dataset, all models generalize greatly with almost similar performance scores to the age diverse dataset. For XceptionNet, the AUC score is 0.9992, for EfficientNet it is 0.9997, and for Lipforensics it is 0.9977. The AUC score of FaceForensics test datasets is higher than the age-diverse test datasets for all three models. For PAUC and EER, these scores are competitive between these datasets.

All these results suggest that the model trained on an age-diverse dataset has higher accuracy scores across all the models and test datasets. Among all models, EfficientNet has outstanding performance with higher scores for AUC, PAUC, and EER metrics across the table. This result shows that the models have strong cross-dataset generalization, validating the effectiveness of the age-diverse dataset in the training process. The performance drop on the Page 97 Celeb-DF dataset suggests a probability that domain-specific features and compression artifacts might be contributing to the lower scores.

\subsubsection{Age-Specific Evaluation}
This evaluation step includes a more detailed analysis of model performance across different age group labels. Here, the model trained with generated age-diverse dataset is evaluated by disaggregating the performance by age group. This analysis is critical for assessing the contribution of final age-diverse dataset in mitigating the fairness issue on deepfake detection.

\begin{table}[ht]
\centering
\caption{Age-specific Evaluation across datasets for models trained with an age-diverse dataset}
\small
\begin{tabular}{|l|l|c|c|c|c|c|c|c|c|c|}
\hline
\multirow{2}{*}{Age Group} & \multirow{2}{*}{Metric} & \multicolumn{3}{c|}{Age-Diverse} & \multicolumn{3}{c|}{Celeb-DF} & \multicolumn{3}{c|}{FaceForensics++} \\
\cline{3-11}
 & & Xec & Eff  & Lip & Xec & Eff & Lip & Xec & Eff & Lip \\
\hline
Overall   & AUC   & 0.9983 & 0.9994 & 0.9970 & 0.9963 & 0.9985 & 0.9927 & 0.9992 & 0.9997 & 0.9977 \\
Overall   & PAUC  & 0.9993 & 0.9998 & 0.9990 & 0.9954 & 0.9983 & 0.9927 & 0.9992 & 0.9997 & 0.9981 \\
Overall   & EER   & 0.0134 & 0.0084 & 0.0213 & 0.0214 & 0.0146 & 0.0365 & 0.0085 & 0.0055 & 0.0202 \\
\hline
0--10     & AUC   & 0.9999 & 0.9999 & 0.9996 & None   & None   & None   & None   & None   & None   \\
0--10     & PAUC  & 1.0000 & 1.0000 & 1.0000 & None   & None   & None   & None   & None   & None   \\
0--10     & EER   & 1.0000 & 1.0000 & 1.0000 & None   & None   & None   & None   & None   & None   \\
\hline
10--18    & AUC   & 0.9998 & 0.9999 & 0.9998 & None   & None   & None   & None   & None   & None   \\
10--18    & PAUC  & 1.0000 & 1.0000 & 1.0000 & None   & None   & None   & None   & None   & None   \\
10--18    & EER   & 0.0064 & 0.0062 & 0.0072 & None   & None   & None   & None   & None   & None   \\
\hline
19--35    & AUC   & 0.9970 & 0.9984 & 0.9918 & 0.9969 & 0.9983 & 0.9907 & 0.9992 & 0.9997 & 0.9979 \\
19--35    & PAUC  & 0.9972 & 0.9985 & 0.9928 & 0.9973 & 0.9983 & 0.9918 & 0.9993 & 0.9997 & 0.9985 \\
19--35    & EER   & 0.0210 & 0.0144 & 0.0386 & 0.0224 & 0.0154 & 0.0407 & 0.0088 & 0.0053 & 0.0192 \\
\hline
36--50    & AUC   & 0.9971 & 0.9994 & 0.9967 & 0.9957 & 0.9988 & 0.9955 & 0.9992 & 0.9997 & 0.9972 \\
36--50    & PAUC  & 0.9976 & 0.9997 & 0.9985 & 0.9912 & 0.9985 & 0.9947 & 0.9987 & 0.9996 & 0.9968 \\
36--50    & EER   & 0.0142 & 0.0097 & 0.0245 & 0.0197 & 0.0131 & 0.0307 & 0.0078 & 0.0063 & 0.0237 \\
\hline
51+       & AUC   & 0.9999 & 0.9999 & 0.9998 & None   & None   & None   & None   & None   & None   \\
51+       & PAUC  & 1.0000 & 1.0000 & 1.0000 & None   & None   & None   & None   & None   & None   \\
51+       & EER   & 0.0041 & 0.0081 & 0.0036 & None   & None   & None   & None   & None   & None   \\
\hline
\end{tabular}
\vspace{0.5em}
\begin{flushleft}
\footnotesize
\textbf{Note:} 
"Xec" = XceptionNet, "Eff" = EfficientNet, "Lip" = LipForensics.  
"None" indicates no available data for that age group in the respective model–dataset combination.
\end{flushleft}
\end{table}

In the Table 5, we can notice there are null values for a couple of cells. The test dataset of Celeb-DF \cite{li2020celebdflargescalechallengingdataset} and FaceForensic++ \cite{rossler2019faceforensics++} does not have data for 0-10, 10-18, and 51+ age groups, resulting in null values.

For the age diverse dataset, the AUC, pAUC, and EER scores across all age groups demonstrate higher accuracy with lower errors. The EfficientNet \cite{Koonce2021} model has the highest value of AUC, pAUC, and EER values, specifically for the 0-10 age group. XceptionNet \cite{chollet2017xceptiondeeplearningdepthwise} has similar performance scores, indicating great performance. For Lipforensics \cite{haliassos2021lipsdontliegeneralisable}, in the age group 36-50, the Page 98 EER score is highest at 0.0245, indicating slightly weaker performance. The model was able to generalize across all age groups in the age-diverse dataset. 

If we only compare the performance for age groups 19-35 and 36-50, the model was able to obtain a high-performance score for Celeb-DF and FaceForensics++ datasets while it was trained with an age-diverse dataset. For FaceForensics++, the AUC values range from 0.9957 to 0.9999, pAUC from 0.9912 to 0.9997, and EER from 0.0053 to 0.0237. Similarly, for Celeb-DFV2, the AUC ranges from 0.9907 to 0.9999, pAUC from 0.9718 to 0.9997, and EER from 0.0131 to 0.0407. For the age groups available in these source datasets, the model shows higher performance. 

To summarize, the performance score of all three models shows high accuracy across all age groups, indicating the effectiveness of the age-diverse dataset used to train the model.

\subsubsection{Cross-Dataset Comparison}
This comparative evaluation is used to assess the influence of an age-diverse dataset on the training of deepfake detection models. it performance was compared with the models trained with the source datset (Celeb-Df \cite{li2020celebdflargescalechallengingdataset}, and FaceForensics++ \cite{rossler2019faceforensics++}). Evaluation tables of AUC, pAUC, and EER metrics are generated for models trained with age-diverse,  Celeb-DF, and FaceForensics++ datasets. The tables (Table 4, Table 6, and Table 7) illustrate the comparative performance of models trained with these three datasets.

\begin{table}[htbp]
\centering
\caption{Evaluation Metrics of deepfake detection model trained with Celeb-DF dataset and tested on age-diverse dataset, Celeb-DF, and Faceforensics++ datasets.}
\begin{tabular}{|c|c|c|c|c|c|c|c|c|c|}
\hline
\textbf{Test Set} & \multicolumn{3}{c|}{\textbf{XceptionNet [celeb]}} & \multicolumn{3}{c|}{\textbf{EfficientNet [Celeb-DF]}} & \multicolumn{3}{c|}{\textbf{LipForensics [celeb]}} \\
\cline{2-10}
& AUC & PAUC & EER & AUC & PAUC & EER & AUC & PAUC & EER \\
\hline
Age-diverse & 0.6428 & 0.8752 & 0.3973 & 0.6288 & 0.8583 & 0.4055 & 0.6513 & 0.8686 & 0.388 \\
\hline
Celeb-DF    & 0.9998 & 0.9998 & 0.0058 & 0.9999 & 0.9999 & 0.0047 & 0.9995 & 0.9995 & 0.01 \\
\hline
FaceForensics++ & 0.5478 & 0.5793 & 0.4627 & 0.5563 & 0.5682 & 0.456 & 0.526 & 0.5479 & 0.4753 \\
\hline
\end{tabular}
\end{table}

\begin{table}[htbp]
\renewcommand{\arraystretch}{1.3}
\centering
\caption{Evaluation Metrics of deepfake detection model trained with FaceForensics++ dataset and tested on age-diverse dataset, Celeb-DF, and FaceForensics++ dataset.}
\begin{tabular}{|c|c|c|c|c|c|c|c|c|c|}
\hline
\textbf{Test Set} & \multicolumn{3}{|c|}{\makecell{\textbf{XceptionNet} \\ \textbf{[FaceForensics++]}}} & \multicolumn{3}{|c|}{\makecell{\textbf{EfficientNet} \\ \textbf{[FaceForensics++]}}} & \multicolumn{3}{|c|}{\makecell{\textbf{LipForensics} \\ \textbf{[FaceForensics++]}}} \\

\hline
 & AUC & PAUC & EER & AUC & PAUC & EER & AUC & PAUC & EER \\
\hline
Balanced & 0.8833 & 0.9609 & 0.1972 & 0.8468 & 0.9487 & 0.2236 & 0.8726 & 0.9569 & 0.2028 \\
\hline
Caleb    & 0.5599 & 0.5818 & 0.4581 & 0.5851 & 0.5949 & 0.4417 & 0.569  & 0.5859 & 0.4467 \\
\hline
FaceForensics++ & 0.9999 & 0.9999 & 0.0024 & 0.9999 & 0.9999 & 0.0041 & 0.9999 & 1.0000 & 0.0029 \\
\hline
\end{tabular}
\end{table}

The comparison of these three tables illustrates that the age-diverse dataset demonstrated strong generalization to unseen datasets (Celeb-DF and FaceForensics++ datasets). The AUC scores for the age-diverse dataset trained model range between (0.9927-0.9997), indicating strong performance score. Also, the pAUC is between (0.9927-0.999), supporting the AUC scores, and the EER has the lower values ranging between (0.0084-0.0365). This demonstrates that the age-diverse dataset enabled models to generalize effectively across different datasets.

In contrast, Celeb-DF and FaceForensics++ datasets has the performance score exceptionally accurate for their dataset; however, it gets significantly lower once validated using the age-diverse dataset and other source datasets. In models trained with the Celeb-DF dataset, the AUC ranges Page 100 between 0.9998-1.0 for its dataset; however, on the age-diverse dataset, the AUC score drops to (0.62-0.65), and for FaceForensics++, it is (0.55-0.52). Similarly, for models trained with FaceForensics++ datasets, the AUC values are approximately 0.9999 for all models when tested with their dataset, whereas for the age-diverse dataset, the AUC value is reduced to (0.88- 0.84), and for Celeb-DF, it is (0.58-0.55).

This strongly suggests that the model trained on FaceForensics++ and Celeb-DF tends to memorize the dataset-specific properties and struggles to detect deepfakes when introduced to a new dataset. On the other hand, the age-diverse dataset was robust against different datasets and age-aware, making it effective at training the deepfake detection models.

\subsection{Findings Summary}

The following list is a summary of all the observations and findings that we discussed in
the above section:
\begin{itemize}
\item The source datasets (Celeb-DF \cite{li2020celebdflargescalechallengingdataset} and FaceForensics++ \cite{rossler2019faceforensics++}) demonstrate skewed age distribution, particularly missing data for age groups such as 0-10, 10-18, and 51+.
\item A total of 2639 synthetic data points were generated using SimSwap to age-diversify the fake
sample across underrepresented age groups. The average SSIM score for synthetic data
was 0.40, and PSNR was 0.28 dB, which indicates acceptable synthetic data generation.
\item The final age-diverse dataset combines the available real and fake datasets
with the created synthetic videos, resulting in an age-diverse dataset for model training with a
total of 6327 data points.
\item All models (XceptionNet \cite{chollet2017xceptiondeeplearningdepthwise}, EfficientNet \cite{Koonce2021}, and LipForensics \cite{haliassos2021lipsdontliegeneralisable}) trained on an age-diverse dataset
achieved higher accuracy scores greater than 0.998 and a lower EER rate less than 0.014.
When validated with other source data, the model's accuracy was reduced by a small
Number, such as for Celeb, the AUC was approximately 0.996, and for FaceForensics, it
was 0.999.
\item When comparing this value with the cross dataset (models trained with Celeb-DF and
FaceForensics dataset), these models performed poorly when tested with the cross-dataset, indicating a strong overfitting issue and poor generalization.
\item Age-specific evaluation illustrated a robust detection across all age groups when trained
on age-diverse data. Also, the model generalized well with the available data for cross-dataset in age groups 18-35 and 36-50. This shows that the age-diverse dataset is capable of
fairer deepfake detection.
\end{itemize}

Overall, the creation of an age-diverse dataset has shown a significant impact on
improving the model’s accuracy on deepfake detection overall and across age demographics. It
also outperformed current datasets on generalizability across cross-dataset performance.

\section{Limitations, Conclusion, and Future Direction}

This study has two primary limitations: (1) The quality of the generated Synthetic videos achieved a moderate score (SSIM ~$\approx$~ 0.40), limited by time constraints; and (2) Only the generated age-diverse dataset and source dataset were utilized for validation, lacking validation on truly unseen or real-world data.

Regardless of limitations, this study was successful at creating an age-diverse deepfake dataset that will help to reduce age bias in deepfake detection datasets. This study identified the gap in age diversity in the source data from Celeb-DF \cite{li2020celebdflargescalechallengingdataset} and FaceForensics++ \cite{rossler2019faceforensics++} datasets.  By combining real and fake videos from source datasets (Celeb-DF and FaceForensics++) with synthetic videos augmented using UTKFace \cite{zhifei2017cvpr} images and source videos, the resulting dataset achieved significantly improved performance across all age groups. 

The comparative analysis shows that the models (XceptionNet, EfficientNet, and LipForensics) trained on the age-diverse dataset outperform those trained on source datasets. The age-diverse dataset was able to demonstrate improved fairness, robust age-specific performance, and strong cross-dataset generalization when evaluated using standard metrics such as AUC, pAUC, and EER. This study contributes a reliable and validated age-diverse deepfake dataset and pipeline that can serve as a foundation for future fairness-aware research in deepfake detection.

Future work can include (1) improvement of the quality of synthetic data for diverse age groups, (2) inclusion of unseen data sources in validation to better assess real-world generalization, and (3) expansion of fairness assessment in additional demographic variables such as gender and ethnicity. Improving the quality of synthetic data for diverse age groups can be beneficial for better identity preservation, reduced artifacts, and improved motion consistency. Inclusion of unseen data will provide a more accurate measure of the model's robustness, reliability, and fairness while detecting deepfakes in unfamiliar styles, identities, and compression
artifacts. Finally, inclusion of additional demographics will enhance the fairness of the age-diverse dataset and further support bias mitigation in deepfake detection.

\section{Acknowledgement}

I would like to express my sincere gratitude to the Grand Canyon University faculty,
Professor Kevin Abreu, Dr. Michele Bonnett, Dr.Mohamad Salen, and Professor Jonathan
Pollyn, for their invaluable guidance, support, and encouragement throughout this project. Their
insights and feedback played a crucial role in shaping the direction and quality of this study. I
am thankful to my fellow students who contributed through discussions, feedback, and reviews,
helping to identify key gaps and inspire them for continual improvement.

Special thanks to the developers and maintainers of open-source tools and datasets such
as Celeb-DF \cite{li2020celebdflargescalechallengingdataset}, FaceForensics++ \cite{rossler2019faceforensics++}, UTKFace \cite{zhifei2017cvpr}, SimSwap \cite{10.1145/3394171.3413630}, InsightFace \cite{deng2018menpo} \cite {guo2021sample} \cite {deng2020subcenter}, and DeepFace \cite{serengil2024lightface},
which were instrumental in creating and evaluating the age-diverse deepfake dataset and the
overall deepfake detection pipeline.

\bibliographystyle{unsrt}

\end{document}